# A SEMI-AUTOMATIC METHOD FOR DOCUMENT CLASSIFICATION IN THE SHIPPING INDUSTRY

**Narayanan Arvind (Yetirajan*)**

**IN-D by Emulya Fintech PTE LTD**

## ABSTRACT

In the shipping industry, document classification plays a crucial role in ensuring that the necessary documents are properly identified and processed for customs clearance. OCR technology is being used to automate the process of document classification, which involves identifying important documents such as Commercial Invoices, Packing Lists, Export/Import Customs Declarations, Bills of Lading, Sea Waybills, Certificates, Air or Rail Waybills, Arrival Notices, Certificate of Origin, Importer Security Filings, and Letters of Credit. By using OCR technology, the shipping industry can improve accuracy and efficiency in document classification and streamline the customs clearance process. The aim of this study is to build a robust document classification system based on keyword frequencies. The research is carried out by analyzing "Contract-Breach" law documents available with IN-D. The documents were collected by scraping the Singapore Government Judiciary website. The database developed has 250 "Contract-Breach" documents. These documents are splitted to generate 200 training documents and 50 test documents. A semi-automatic approach is used to select keyword vectors for document classification. The accuracy of the reported model is 92.00 %.

## NOMENCLATURE

| | |
|---|---|
| a | Corpus feature vector |
| b | Document feature vector |
| 1 | Contract-Breach class |
| 0 | Others class |

## INTRODUCTION

Document classification is an essential task in the shipping industry as it involves processing and managing vast amounts of information related to shipping operations. To ensure compliance with regulations and to enhance efficiency, shipping companies need to classify various documents such as bills of lading, cargo manifests, and customs declarations accurately [1]. With the advent of digitalization, these documents are increasingly being submitted in a digital format, which necessitates the use of automated techniques for document classification. However, the accuracy of automated classification methods can be affected by various factors such as the quality of the document image and variations in document structure.

Text-based document classification methods are widely used in the shipping industry to automate the process of identifying and categorizing various types of documents. These methods typically involve the use of natural language processing (NLP) techniques to extract features and patterns from the text in the documents. One popular approach is the use of

Corresponding author, Email: arvind.narayanan@in-d.ai Tel: +91-8349232657
M.Tech. Software Engineering (WILP), Birla Institute of Technology and Science, Pilani
Fellow AI/ML, GradValley DataScience, Coimbatore
B.Tech. / M.Tech. Dual degree Ocean Engineering and Naval Architecture, IIT Kharagpur
*meaning king of the ascetics*

machine learning algorithms such as decision trees [2], support vector machines [3], and neural networks [4] to classify documents based on their textual content. Another approach involves the use of rule-based systems [5], where predefined rules are used to classify documents based on their content and metadata. Additionally, topic modeling techniques such as Latent Dirichlet Allocation (LDA) [6] can be used to identify the main topics and themes present in a set of documents. While each of these methods has its own advantages and limitations, the choice of the most suitable method depends on the specific requirements and constraints of the shipping industry context.

Deep learning methods have become increasingly popular for document classification in recent years due to their ability to learn complex patterns and features in data. One such method is Convolutional Neural Networks (CNNs) [7], which are often used in image classification but can also be applied to text classification tasks by treating text as a sequence of one-dimensional signals. CNNs can automatically learn relevant features and patterns from raw text data, which makes them effective in classifying documents. Another deep learning method is Recurrent Neural Networks (RNNs) [8], which are designed to process sequences of variable length. RNNs are particularly useful for document classification tasks that require understanding the context of words within a document, such as sentiment analysis or topic classification. Long Short-Term Memory (LSTM) [9] networks are a type of RNN that can effectively capture long-term dependencies in text data. In addition to CNNs and RNNs, there are other deep learning methods that can be used for document classification, such as Transformers [10] and BERT [11] (Bidirectional Encoder Representations from Transformers), which are based on self-attention mechanisms and have achieved state-of-the-art results in many NLP tasks, including document classification. These models can be pre-trained on large amounts of data and then fine-tuned on specific document classification tasks, allowing them to generalize well and achieve high accuracy. Robert P. et.al. [12] present an automated solution to document classification into hierarchies by using wikipedia documents, their document feature vectors and their associated hierarchical topic tags.

In this research paper, we propose a semi-automatic method for document classification in the shipping industry, which leverages machine learning and user input to achieve high accuracy and efficiency.

## A SEMI-AUTOMATIC METHOD FOR DOCUMENT CLASSIFICATION

*2.1 Preprocessing*

We work with 250 "Contract-Breach" samples and 50 "Criminal-Law" samples available with IN-D. The dataset was obtained by scraping the Singapore Government Judiciary website [13]. Only PDF documents were considered in this study namely *.pdf and *.PDF document types. The model can be easily extended to other document types by using suitable document conversion and OCR techniques. We start by the extraction of text from the documents using the open-source PDFMiner API [14].

*2.2 Solution methodology*

We find the corpus feature vector for the "Contract-Breach" training documents by humanly selecting top 50 keywords from 50 "Contract-Breach" samples. These 50 keywords are selected based on human judgment and experience and hence can be advantageous in document classification. These 50 keywords and their frequencies are then found in the 200 "Contract-Breach" training samples. The top 20 keywords are selected based on their word counts. This vector of 20 keywords and their counts forms the corpus feature vector for "Contract-Breach" samples.

These 20 keywords and their word counts are now found in the test document to form the document feature vector. The document feature vector and the corpus feature vector are used to find the cosine similarity for the test document with the corpus. The test document is then classified into a target category based on a suitable similarity threshold.

$$sim = \frac{ab}{|a||b|} \quad (1)$$
$$class = 0 \; if \; sim < threshold \; else \; 1 \quad (2)$$

Where **a** and **b** are the corpus and document feature vectors respectively. **0** represents "Others" class while **1** represents "Contract-Breach" class.

## EXPERIMENTS

In this section we propose the methodology for carrying out our experimental studies. We use 250 "Contract-Breach" documents available with IN-D. Python3 is the chosen language for programming. Numpy [15], Pandas [16] and PDFMiner OCR engine are the major open-source libraries used.

*3.1 Dataset*

The dataset available with IN-D consists of 250 "Contract-Breach" samples scraped from the Singapore Government Judiciary website. 200 "Contract-Breach" samples were used to train the semi-automatic classifier while 50 samples were used for testing. In addition, 50 "Criminal-Law" samples were scraped from the Singapore Government Judiciary website to test the model performance on "Others" (non "Contract-Breach") category. The samples range from approximately 3 pages to 150 pages in length.

*3.2 Experimental protocol*

We start by finding the top 50 keywords humanly from 50 "Contract-Breach" training samples. This is advantageous as humans can bring their experience and discerning capabilities to select relevant keywords. These 50 keywords are then found in the 200 training samples along with their word counts. The top 20 keywords, as per their frequencies, are selected for modeling.

These 20 keywords along with their frequencies in the training data form the corpus feature vector. The 20 keywords are then found in the test document to form the document feature vector. The document feature vector and the corpus feature vector are then used to find the cosine similarity score for the given test document. The test document is then classified as a "Contract-Breach" document or "Others" category document based on a threshold similarity score. The mathematical formulations for the solution can be found in equations (1) and (2).

*3.3 Results and discussions*

The following top 20 keywords were found for the "Contract-Breach" class using the semi-automatic method.

Top20= ['Breach','Contract','Owe','Sum','Cost','Agreement', 'Pay','MOU','Rent','Dispute','Amount','Damage', 'Obligation','Liability','Document','Differential','Material','Approval','Loss', 'Offer']

Table 1. Top 20 keywords for the "Contract-Breach" class.

Our model achieves an accuracy of 92% on the classification of "Contract-Breach" documents using a similarity threshold of 0.6 (Table 2). This means, 92% of "Contract-Breach" documents are correctly classified when a threshold of 0.6 is used for the cosine similarity metric. Confusion matrices for the classifier are presented at thresholds of 0.6 and 0.65 (Table 3 and Table 4). Accuracy metrics for the model are also presented in Table 5. As can be seen from the results, the model performs best at a threshold of 0.6 for the cosine similarity metric.

| Snum | Threshold | Num test_docs | Accuracy |
|---|---|---|---|
| 1 | 0.6 | 50 | 92 % |
| 2 | 0.65 | 50 | 78 % |
| 3 | 0.7 | 50 | 62 % |
| 4 | 0.75 | 50 | 44 % |
| 5 | 0.8 | 50 | 30 % |
| 6 | 0.85 | 50 | 18 % |
| 7 | 0.9 | 50 | 4 % |

Table 2. (Above) accuracy at various thresholds for the IN-D "Contract-Breach" test dataset.

| Threshold=0.6 | | Predicted | |
|---|---|---|---|
| | | Contract-Breach | Others |
| Actual | Contract-Breach | 46 | 4 |
| | Others | 2 | 48 |

Table 3. (Above) confusion matrix for the classifier at threshold=0.6

| Threshold=0.65 | | Predicted | |
|---|---|---|---|
| | | Contract-Breach | Others |
| Actual | Contract-Breach | 39 | 11 |
| | Others | 1 | 49 |

Table 4. (Above) confusion matrix for the classifier at threshold=0.65

| Threshold | Precision | Recall | F1 Score | Accuracy |
|---|---|---|---|---|
| 0.6 | 0.96 | 0.92 | 0.94 | 94 |
| 0.65 | 0.98 | 0.78 | 0.87 | 88 |

Table 5. (Left) accuracy metrics for the classifier at threshold = 0.6 and threshold = 0.65

## CONCLUSIONS

A robust semi-automatic method for document classification was presented for Intelligent Document Processing (IDP) in the shipping industry in this paper. "Contract-Breach" documents and "Criminal-Law" documents available with IN-D, were used for our experiments. A corpus vector and a document vector are extracted for the corpus and the test document respectively, which are then used for document classification using cosine similarity. Our model uses a semi-automatic approach to achieve high document classification accuracy. The method can be extended to include more document categories for classification.

## ACKNOWLEDGEMENTS

This work has been supported by the R&D labs at IN-D by Emulya Technologies Pvt. Ltd. The OCR engine used is PDFMiner. The author would especially like to thank OpenAI and their recently (2022-23) released ChatGPT AI software for aiding in writing text for this document. The author would also like to thank the open source community for providing various libraries to complete this project. The author also thanks Robert P. et.al. [12] for inspiring the research presented in this paper.